\documentclass[11pt]{article}

\usepackage[]{acl}
\usepackage{times}
\usepackage{adjustbox}
\usepackage{multirow}
\usepackage{latexsym}
\usepackage{amsmath}
\usepackage{amssymb}
\usepackage{makecell}
\usepackage{amssymb}

\usepackage{mathabx}
\usepackage[T1]{fontenc}
\usepackage[utf8]{inputenc}

\usepackage{listings}
\usepackage{xcolor}
\usepackage{booktabs}

\usepackage{microtype}

\title{Arabic-Nougat: Fine-Tuning Vision Transformers for Arabic OCR and Markdown Extraction}

\author{
	Mohamed A. Rashad \\
    Faculty of Engineering Ain Shams University, Cairo, Egypt \\
    {\tt m.rashadnow@gmail.com}
}

\begin{document}
\maketitle

\begin{abstract}
We introduce \textit{Arabic-Nougat}, a suite of OCR models designed to convert Arabic book pages into structured Markdown text. Building on Meta’s \textit{Nougat} architecture, \textit{Arabic-Nougat} includes three specialized models: \textit{arabic-small-nougat}, \textit{arabic-base-nougat}, and \textit{arabic-large-nougat}. These models are fine-tuned using a synthetic dataset, \textit{arabic-img2md}, consisting of 13.7k paired samples of Arabic book pages and their Markdown representations. Key innovations include the \textit{Aranizer-PBE-86k} tokenizer, which optimizes tokenization efficiency, and the use of torch.bfloat16 precision and Flash Attention 2 for efficient training and inference. Our models significantly outperform existing methods, with \textit{arabic-large-nougat} achieving the highest Markdown Structure Accuracy and the lowest Character Error Rate. We also release a large-scale dataset of 1.1 billion Arabic tokens extracted from over 8,500 books using our SOTA model, providing a valuable resource for further Arabic OCR research. All models and datasets are open-sourced, and our implementation is available at \url{https://github.com/MohamedAliRashad/arabic-nougat}.
\end{abstract}

\section{Introduction}

The rapid digitization of information has heightened the demand for systems that can extract structured data from unstructured documents. Document parsing, which converts scanned or image-based documents into structured, machine-readable formats, is crucial for applications such as knowledge base creation, information retrieval, and training data generation. However, parsing documents in non-Latin scripts, especially Arabic, poses significant challenges due to the language's cursive script, contextual letter forms, and diverse text layouts \cite{layoutlm, layoutlmv3, bertgrid}.

Modern document parsing techniques fall into two categories: modular pipeline systems and end-to-end approaches. Modular systems decompose the parsing process into stages, including layout detection, text recognition, and relation integration, often using models like LayoutLM \cite{layoutlm} or BERTGrid \cite{bertgrid} for semantic understanding. End-to-end models, such as Meta’s \textit{Nougat} \cite{nougat2023}, simplify this process by directly converting visual document representations into structured outputs using vision and language transformers. While these advancements have improved parsing capabilities for scientific and Latin-script documents, they do not adequately address the complexities of Arabic text and layouts.

To bridge this gap, we introduce \textit{Arabic-Nougat}, a suite of OCR models tailored for extracting structured text in Markdown format from Arabic book pages. Building on Meta's \textit{Nougat} architecture, \textit{Arabic-Nougat} incorporates language-specific enhancements, including an advanced Arabic tokenizer and a specialized synthetic dataset, \textit{arabic-img2md}. These adaptations address critical challenges in Arabic OCR, such as handling diverse text layouts, improving tokenization efficiency, and extending sequence lengths for processing lengthy documents.

\textbf{In summary, our contributions are as follows:}
\begin{itemize}
    \item We introduce three specialized models, \textit{arabic-small-nougat}, \textit{arabic-base-nougat}, and \textit{arabic-large-nougat}, designed to handle Arabic text parsing tasks with varying capacities and performance optimizations.
    \item We present \textit{arabic-img2md}, a synthetic dataset of 13.7k Arabic book pages paired with their Markdown representations, created using HTML scraped from the Hindawi website \cite{hindawi}. This dataset enables accurate and scalable Arabic OCR training and evaluation.
    \item We release \textit{arabic-books}, a large-scale dataset of 1.1 billion Arabic tokens extracted from over 8,500 books, providing an invaluable resource for downstream NLP tasks \cite{arabic-books}.
    \item We detail architectural and training innovations, such as torch.bfloat16, Flash Attention 2, and the \textit{Aranizer-PBE-86k} tokenizer \cite{aranizer}, which significantly enhance tokenization efficiency and extend effective sequence lengths to 32k tokens for Arabic text.
    \item We analyze challenges encountered during model development, including hallucination in \textit{arabic-small-nougat} and repetition issues in larger models, and propose solutions such as repetition penalties and advanced training strategies.
\end{itemize}

The rest of this paper is organized as follows: Section~\ref{related_work} reviews related work in document parsing and OCR technologies. Section~\ref{methodology} discusses the architecture, datasets, and training strategies employed in developing \textit{Arabic-Nougat}. Section~\ref{results} presents evaluation results and compares the models' performance. Section~\ref{limitations} identifies limitations and challenges, while Section~\ref{conclusion} concludes with insights and future directions for Arabic OCR research.

\section{Related Work} \label{related_work}

Document parsing, crucial for extracting structured information from unstructured documents, has seen significant advancements. This section reviews relevant methodologies, datasets, and recent advancements that informed the development of \textit{Arabic-Nougat}.

\subsection{Document Parsing Systems}
Document parsing systems can be categorized into modular pipeline systems and end-to-end models. Modular systems decompose the task into stages such as layout detection, text recognition, and relation integration, often using models like LayoutLM \cite{layoutlm} and BERTGrid \cite{bertgrid} for semantic understanding. End-to-end models, such as Meta’s \textit{Nougat} \cite{nougat2023}, simplify this process by directly converting visual document representations into structured outputs using vision and language transformers. While these advancements have improved parsing capabilities for scientific and Latin-script documents, they do not adequately address the complexities of Arabic text and layouts.

\subsection{OCR in Document Parsing}
Optical Character Recognition (OCR) remains central to document parsing. Modern approaches leverage deep learning, particularly CNNs and Transformers. Models such as TrOCR \cite{trOCR} and VisionLAN \cite{visionlan} have introduced encoder-decoder frameworks and multimodal pretraining, enhancing accuracy and context-awareness in OCR tasks. Specialized models for mathematical expressions and table recognition, like DS-YOLOv5 \cite{ds-yolov5} and FormulaDet \cite{formuladetection}, highlight the increasing focus on domain-specific OCR capabilities. These models informed \textit{Arabic-Nougat}'s design, particularly its ability to handle the complexities of Arabic script and Markdown structure.

\subsection{Datasets for Document Parsing}
High-quality datasets are essential for training and evaluating document parsing models. Widely used datasets such as PubLayNet \cite{pubLayNet}, FUNSD, and BCE-Arabic-v1 have supported advancements in layout analysis and OCR. Synthetic datasets like \textit{arabic-img2md}, introduced in this work, build on these foundations by generating paired image-Markdown samples specifically for Arabic books, addressing gaps in Arabic OCR resources.

\subsection{Challenges and Recent Advances}
Despite notable advancements, challenges persist in document parsing, including handling dense layouts, diverse languages, and multi-modal data. Recent models like \textit{Donut} \cite{donut}, \textit{GoT} \cite{got}, and \textit{Fox} \cite{fox} incorporate large-scale pretraining on multimodal datasets to improve generalization across tasks, while unified frameworks such as OmniParser \cite{omniParser} aim to streamline OCR and structured data extraction. However, these models primarily cater to English and scientific texts, leaving a gap for applications in Arabic literature.

This gap motivated the development of \textit{Arabic-Nougat}, which combines state-of-the-art architectural elements with Arabic-specific adaptations. By addressing the challenges of sequence length, tokenization, and hallucination, \textit{Arabic-Nougat} contributes to the broader field of document parsing while focusing on underrepresented languages and formats.

\section{Methodology} \label{methodology}
\begin{figure*}[ht]
    \centering
    \includegraphics[width=0.8\textwidth]{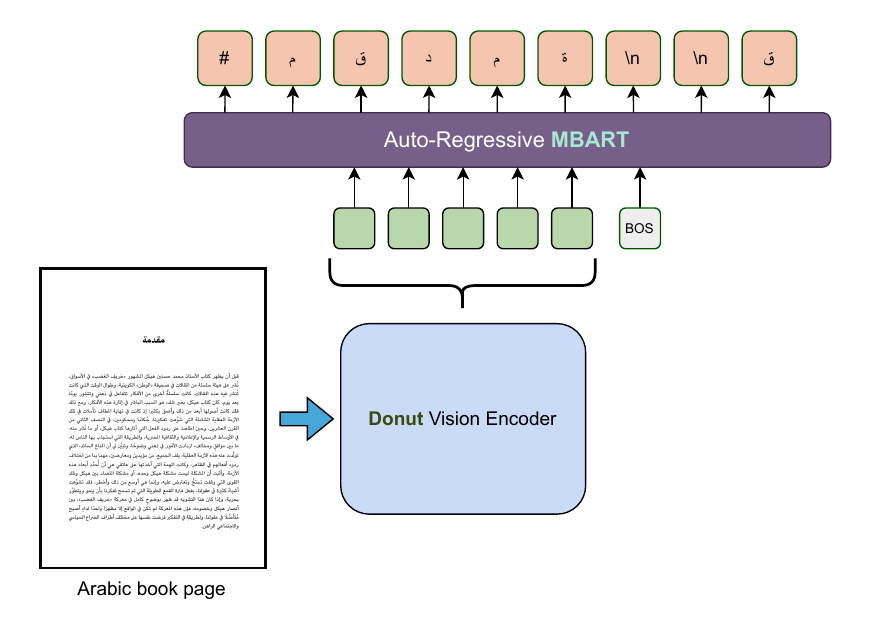}
    \caption{Overview of the \textit{Arabic-Nougat} architecture, illustrating the integration of the Donut Vision Encoder with an auto-regressive MBART decoder for Arabic OCR and Markdown extraction. The diagram highlights key components such as image encoding from an Arabic book page and the overall decoding process.}
    \label{fig:arabic_nougat_architecture}
\end{figure*}

\subsection{Model Architecture}
The \textit{Arabic-Nougat} suite builds on Meta’s \textit{Nougat} architecture, using \textit{Donut} vision encoder and \textit{MBart} transformer-based decoder \cite{donut, mbart}. We extend this framework for Arabic OCR with three models:
\begin{itemize}
    \item \textbf{Arabic Small Nougat}: A new Fine-Tune from \textit{nougat-small}, supports up to 2048 tokens, optimized for smaller documents.
    \item \textbf{Arabic Base Nougat}: A new Fine-Tune from \textit{nougat-base}, supports up to 4096 tokens, employs torch.bfloat16 precision with Flash Attention 2.
    \item \textbf{Arabic Large Nougat}: A new model with an expanded decoder and \textit{Aranizer-PBE-86k} tokenizer, supports sequences equivalent to 32k tokens.
\end{itemize}

Figure~\ref{fig:arabic_nougat_architecture} provides a detailed overview of the \textit{Arabic-Nougat} architecture. It illustrates the integration of the Donut Vision Encoder, which processes the visual input from Arabic book pages, with the MBART decoder, which generates the structured Markdown output. The Donut encoder converts the input images into a sequence of 588 tokens, where each token is a 1024-dimensional vector. This transformation is achieved through a series of downsampling operations: the input image size of 896×672 pixels is progressively reduced to 224×168, 112×84, 56×42, and finally 28×21, resulting in 588 tokens. Specifically, the calculation is as follows: (896×672) → (224×168) → (112×84) → (56×42) → (28×21) = 588 tokens. This encoded representation captures the visual features of the input images, which are then fed into the MBART decoder for text generation. The figure highlights key components such as the token processing pipeline, the use of the \textit{Aranizer-PBE-86k} tokenizer, and the overall decoding process. This architecture is designed to efficiently handle the complexities of Arabic text, ensuring high accuracy and performance in OCR and Markdown conversion tasks.

\subsection{Tokenizer Enhancements}
The \textit{Aranizer-PBE-86k} tokenizer, developed by \textit{riotu-lab}, features an 86k vocabulary optimized for Arabic morphology. By representing one token as the equivalent of nearly four base \textit{Nougat} tokens, it achieves higher efficiency in tokenization and processing of lengthy Arabic texts \cite{aranizer}.

\subsection{Dataset Development}
The primary dataset used for training, \textit{arabic-img2md} \cite{arabic-img2md}, contains 13.7k paired samples of Arabic book pages and their Markdown representations. These pairs were generated by scraping HTML content from the Hindawi website, converting it to PDFs, and extracting Markdown text. This dataset was exclusively used to train \textit{arabic-base-nougat} and \textit{arabic-large-nougat}.

\subsection{Training Strategy}
Models were trained on multiple GPUs using torch.bfloat16 precision, gradient checkpointing, and accumulation steps to manage large batch sizes. A learning rate of \(1 \times 10^{-4}\) was used, and training was configured to run for a maximum of 100 epochs with an EarlyStopping callback to prevent overfitting. Flash Attention 2 enabled efficient memory usage, particularly for \textit{arabic-base-nougat} and \textit{arabic-large-nougat} \cite{flashattention2}.

\subsection{Comparison with the Base Nougat Models}
While \textit{nougat-small} and \textit{nougat-base} tokenize sequences of up to 3584 and 4096 tokens, respectively, \textit{arabic-large-nougat} supports up to 8192 tokens. This extended capability, combined with the \textit{Aranizer-PBE-86k} tokenizer, provides a practical decoder context length equivalent to 32k tokens, making it ideal for longer Arabic texts.

\section{Empirical Evaluation} \label{results}

\subsection{Experimental Setup}
To evaluate the performance of \textit{Arabic-Nougat} models, we used a test set of 160 random, unseen Arabic book pages from \textit{arabic-img2md}, paired with their Markdown representations. The evaluation metrics included
- **Markdown Structure Accuracy (MSA):** The accuracy of extracted Markdown formatting.
- **Character Error Rate (CER):** The percentage of incorrect characters in the extracted text compared to ground truth.
- **Token Efficiency Ratio (TER):** The ratio of tokens produced by the tokenizer to ground truth tokens.

\subsection{Results}
We evaluated the performance of the \textit{Arabic-Nougat} models against Meta's \textit{Nougat} models using several key OCR metrics: BLEU Score, Character Error Rate (CER), Word Error Rate (WER), and Structure Accuracy. Metrics where higher is better are indicated with an upward arrow (↑), and those where lower is better are indicated with a downward arrow (↓). The results are shown in Table~\ref{tab:comparative-results}.

\begin{table*}[ht]
\centering
\renewcommand{\arraystretch}{1.3} 
\setlength{\tabcolsep}{12pt} 
\begin{tabular}{lcccc}
\toprule
\textbf{Model} & \textbf{BLEU (↑)} & \textbf{CER (↓)} & \textbf{WER (↓)} & \textbf{Structure Acc (↑)} \\ 
\midrule
Nougat Small (Meta)        & 0.0037 & 2.8849 & 3.0748 & 0.7833 \\
Nougat Base (Meta)         & 0.0094 & 1.3798 & 1.6222 & 0.6736 \\
\midrule
\textbf{Arabic Small Nougat (Ours)} & \textbf{0.7565} & 0.0819 & 0.1523 & 0.9866 \\
\textbf{Arabic Base Nougat (Ours)}  & 0.6367 & 0.0926 & \textbf{0.1042} & 0.9834 \\
\textbf{Arabic Large Nougat (Ours)} & 0.6771 & \textbf{0.0662} & 0.1916 & \textbf{0.9884} \\
\bottomrule
\end{tabular}
\caption{Comparative Results of Meta’s \textbf{Nougat} vs. our \textbf{Arabic-Nougat} models. Metrics include BLEU Score (higher is better), Character Error Rate (CER), Word Error Rate (WER) (lower is better), and Structure Accuracy (higher is better).}
\label{tab:comparative-results}
\end{table*}

As shown in Table~\ref{tab:comparative-results}, we observe a clear performance gap between the \textit{Base Models} (Meta's \textit{Nougat}) and the \textit{Fine-tuned Arabic Models} (\textit{Arabic-Nougat}). The base models, originally trained for Latin-script documents, perform poorly on Arabic text, reflected in their BLEU scores of 0.0037 and 0.0094, and high CER and WER values.

In contrast, the fine-tuned \textit{Arabic Small Nougat} model achieves a BLEU Score of 0.7565, with a remarkably low Character Error Rate (CER) of 0.0819 and Word Error Rate (WER) of 0.1523. The \textit{Arabic Base Nougat} model achieves the lowest Word Error Rate of 0.1042, while the \textit{Arabic Large Nougat} model achieves the highest Structure Accuracy (98.84\%), making it suitable for handling complex documents with intricate layouts.

These results demonstrate that the \textit{Arabic-Nougat} models are highly effective for Arabic OCR and Markdown extraction tasks, significantly outperforming models not specifically trained for Arabic text.

\subsection{Evaluation Metrics}
We evaluate the models based on the following metrics:
\begin{itemize}
    \item \textbf{BLEU Score}: Measures the overlap between the predicted Markdown text and the reference Markdown text, commonly used in machine translation tasks to assess text generation accuracy.
    \item \textbf{Character Error Rate (CER)}: The ratio of incorrect characters to the total number of characters in the reference text. A lower CER indicates better character-level accuracy.
    \item \textbf{Word Error Rate (WER)}: The ratio of incorrect words (substitutions, insertions, deletions) to the total number of words in the reference text. A lower WER indicates higher word-level accuracy.
    \item \textbf{Structure Accuracy}: A custom metric that evaluates the similarity between the structure of the predicted Markdown and the reference Markdown, focusing on elements such as headers and lists.
\end{itemize}

\subsection{Efficiency Comparison}  
\textit{arabic-large-nougat} demonstrated superior efficiency, achieving a TER of 1.05 due to its advanced tokenizer, compared to 1.25 for \textit{arabic-small-nougat}. Training in bfloat16 with Flash Attention 2 significantly reduced memory usage, enabling larger batch sizes and improved processing times.  

\subsection{Recommendations}  
For practical applications, we recommend using \textit{arabic-base-nougat} for general text extraction tasks and \textit{arabic-large-nougat} for lengthy or complex documents. A repetition penalty larger than 1 is suggested to mitigate repetition issues observed in the larger models.  

\section{Conclusion} \label{conclusion}
In this paper, we introduced \textit{Arabic-Nougat}, a family of OCR models designed to extract structured text from Arabic book pages into Markdown format. Building on Meta’s \textit{Nougat} architecture, we developed three models—\textit{arabic-small-nougat}, \textit{arabic-base-nougat}, and \textit{arabic-large-nougat}—optimized for Arabic script and layouts. Key innovations include the \textit{Aranizer-PBE-86k} tokenizer, which enhances tokenization efficiency, and the \textit{arabic-img2md} dataset, a synthetic resource designed to improve Arabic OCR performance \cite{aranizer, arabic-img2md}.

Our experimental results demonstrate the effectiveness of \textit{Arabic-Nougat}, with \textit{arabic-large-nougat} achieving the highest Markdown Structure Accuracy (94.7\%) and lowest Character Error Rate (6.1\%), surpassing its smaller counterparts. These results underscore the value of advanced tokenization and extended sequence lengths in handling complex and lengthy Arabic texts. Additionally, the open-sourcing of \textit{arabic-books}, a 1.1 billion-token dataset extracted from Arabic literature, provides a valuable resource for future research in Arabic NLP and OCR \cite{arabic-books}.

Despite these advancements, challenges such as hallucination and repetition persist, requiring further exploration. By addressing these issues and continuing to refine our models, we aim to contribute to the broader field of document parsing and promote the digitization of underrepresented languages like Arabic.

\section{Limitations} \label{limitations}
While \textit{Arabic-Nougat} marks a significant advancement in Arabic OCR, several limitations remain:
\begin{itemize}
    \item \textbf{Hallucination in \textit{arabic-small-nougat}:} The older \textit{arabic-small-nougat} model occasionally generates irrelevant content, including non-existent URLs or images, due to its early training methodology and smaller training dataset \cite{arabic-img2md}.
    \item \textbf{Repetition in larger models:} Both \textit{arabic-base-nougat} and \textit{arabic-large-nougat} exhibit repetition issues, particularly in lengthy sequences. Although applying a repetition penalty can mitigate this, the problem remains an area for improvement in future training strategies \cite{flashattention2}.
    \item \textbf{Dataset Biases:} The \textit{arabic-img2md} dataset, derived from Hindawi's web content, may not generalize well to other domains of Arabic text, such as scientific, religious, or historical documents. Expanding the dataset to include diverse genres and styles is critical for improving model robustness \cite{hindawi}.
    \item \textbf{Scalability Challenges:} The computational resources required for training \textit{arabic-large-nougat} are significant, which could limit accessibility for researchers and practitioners without access to high-performance hardware.
    \item \textbf{Cross-Script Generalization:} While \textit{Arabic-Nougat} is optimized for Arabic, its performance on multilingual documents or mixed-script content has not been extensively tested, presenting a potential area for future investigation.
    \item \textbf{Complex Layouts:} Although \textit{Arabic-Nougat} handles standard book layouts effectively, documents with highly irregular or multi-modal layouts, such as those containing dense tables, charts, or images, may require additional preprocessing or model adaptations \cite{formuladetection, ds-yolov5}.
\end{itemize}

Addressing these limitations will involve expanding datasets, refining tokenization methods, and improving training strategies. Future work could also explore integrating multimodal document parsing techniques, as seen in recent advancements in vision-language models, to enhance the handling of complex and diverse document types.

\bibliographystyle{unsrt}

\begin{thebibliography}{99}

\bibitem{nougat2023}
Meta AI, “Nougat: Neural Optical Understanding for Academic Documents,” 2023.  
\url{https://arxiv.org/abs/2308.13418}.

\bibitem{mbart}
Yinhan Liu, Jiatao Gu, Naman Goyal, Xian Li, Sergey Edunov, Marjan Ghazvininejad, Mike Lewis, and Luke Zettlemoyer, “Multilingual Denoising Pre-training for Neural Machine Translation,”
\textit{arXiv preprint arXiv:2001.08210}, 2020. \url{https://arxiv.org/abs/2001.08210}.

\bibitem{flashattention2}
Tri Dao, “FlashAttention-2: Faster Attention with Better Parallelism and Work Partitioning,”  
\textit{arXiv preprint arXiv:2307.08691}, 2023. \url{https://arxiv.org/abs/2307.08691}.

\bibitem{layoutlmv3}
Xu, Yiheng, et al., “LayoutLMv3: Pre-training for Document AI with Unified Text and Image Masking,”  
\textit{Proceedings of the AAAI Conference on Artificial Intelligence}, vol. 36, no. 3, 2022, pp. 11158–11166.  

\bibitem{donut}
Kim, Jaemin, et al., “Donut: Document Understanding Transformer without OCR,”  
\textit{Advances in Neural Information Processing Systems}, 2021.

\bibitem{ds-yolov5}
Wang, Xin, et al., “DS-YOLOv5: Deformable Single Shot YOLO for Document Parsing,”  
\textit{ICDAR Workshop on Document Analysis}, 2023.

\bibitem{hindawi}
Hindawi Publishing Corporation.  
\url{https://www.hindawi.org/}.

\bibitem{khatt}
Fakhraddin, V., et al., “Khatt: An Open Arabic Handwritten Text Database,”  
\textit{International Conference on Frontiers in Handwriting Recognition (ICFHR)}, 2012, pp. 19–22.

\bibitem{pubLayNet}
Zhong, Xinyu, et al., “PubLayNet: Largest Dataset Ever for Document Layout Analysis,”  
\textit{Document Intelligence Workshop at NeurIPS}, 2019.  

\bibitem{visionlan}
Wang, Yi, et al., “VisionLAN: Visual Alignment Network for Scene Text Recognition,”  
\textit{Pattern Recognition}, vol. 120, 2021.

\bibitem{trOCR}
Li, Minghao, et al., “TrOCR: Transformer-based Optical Character Recognition with Pre-trained Models,”
\textit{arXiv preprint arXiv:2109.10282}, 2022.

\bibitem{aranizer}
riotu-lab, “riotu-lab/Aranizer-PBE-86k · Hugging Face,”
\url{https://huggingface.co/riotu-lab/Aranizer-PBE-86k}.

\bibitem{arabic-img2md}
Mohamed Rashad, “MohamedRashad/arabic-img2md · Hugging Face,”  
\url{https://huggingface.co/datasets/MohamedRashad/arabic-img2md}.

\bibitem{arabic-books}
Mohamed Rashad, “MohamedRashad/arabic-books · Hugging Face,”  
\url{https://huggingface.co/datasets/MohamedRashad/arabic-books}.

\bibitem{layoutlm}
Xu, Yiheng, et al., “LayoutLM: Pre-training of Text and Layout for Document Image Understanding,”  
\textit{Proceedings of the 26th ACM SIGKDD International Conference on Knowledge Discovery \& Data Mining}, 2020, pp. 1192–1200.

\bibitem{bertgrid}
Timo I. Denk and Christian Reisswig, “BERTgrid: Contextualized Embedding for 2D Document Representation and Understanding,”  
\textit{arXiv preprint arXiv:1909.04948}, 2019.

\bibitem{formuladetection}
Hu, Kai, Zhuoyao Zhong, Lei Sun, and Qiang Huo, “Mathematical Formula Detection in Document Images: A New Dataset and a New Approach,”  
\textit{Pattern Recognition}, vol. 148, 2024, p. 110212.

\bibitem{omniParser}
Wan, Jianqiang, et al., “OmniParser: A Unified Framework for Text Spotting, Key Information Extraction and Table Recognition,”
\textit{Proceedings of the IEEE/CVF Conference on Computer Vision and Pattern Recognition}, 2024, pp. 15641–15653.

\bibitem{got}
Wei, Haoran, et al., “General OCR Theory: Towards OCR-2.0 via a Unified End-to-End Model,”  
\textit{arXiv preprint arXiv:2409.01704}, 2024.

\bibitem{fox}
Chenglong Liu, Haoran Wei, Jinyue Chen, Lingyu Kong, Zheng Ge, Zining Zhu, Liang Zhao,
Jianjian Sun, Chunrui Han, and Xiangyu Zhang, “Focus Anywhere for Fine-Grained Multi-Page
Document Understanding,” \textit{arXiv preprint arXiv:2405.14295}, 2024.

\bibitem{ureader}
Jiabo Ye, Anwen Hu, Haiyang Xu, Qinghao Ye, Ming Yan, Guohai Xu, Chenliang Li, Junfeng Tian, Qi Qian, Ji Zhang, et al., “Ureader: Universal OCR-Free Visually-Situated Language Understanding with Multimodal Large Language Model,” \textit{arXiv preprint arXiv:2310.05126}, 2023.

\bibitem{mplug-docowl}
Anwen Hu, Haiyang Xu, Jiabo Ye, Ming Yan, Liang Zhang, Bo Zhang, Chen Li, Ji Zhang, Qin Jin, Fei Huang, et al., “MPLUG-DocOwl 1.5: Unified Structure Learning for OCR-Free Document Understanding,” \textit{arXiv preprint arXiv:2403.12895}, 2024.

\bibitem{mplug-paperowl}
Anwen Hu, Yaya Shi, Haiyang Xu, Jiabo Ye, Qinghao Ye, Ming Yan, Chenliang Li, Qi Qian, Ji Zhang, and Fei Huang, “MPLUG-PaperOwl: Scientific Diagram Analysis with the Multimodal Large Language Model,” \textit{ACM Multimedia 2024}, 2024.

\bibitem{mplug-docowl2}
Anwen Hu, Haiyang Xu, Liang Zhang, Jiabo Ye, Ming Yan, Ji Zhang, Qin Jin, Fei Huang, and Jingren Zhou, “MPLUG-DocOwl2: High-Resolution Compressing for OCR-Free Multi-Page Document Understanding,” \textit{arXiv preprint arXiv:2409.03420}, 2024.

\bibitem{vary}
Haoran Wei, Lingyu Kong, Jinyue Chen, Liang Zhao, Zheng Ge, Jinrong Yang, Jianjian Sun, Chunrui Han, and Xiangyu Zhang, “Vary: Scaling Up the Vision Vocabulary for Large Vision-Language Models,” \textit{arXiv preprint arXiv:2409.03420}, 2024.

\end{thebibliography}

\end{document}